\documentclass[letterpaper, 10 pt, conference]{ieeeconf}  

\IEEEoverridecommandlockouts                              %
\usepackage{graphicx}
\usepackage{subcaption}
\usepackage{amsmath}
\usepackage{longtable}
\usepackage{tabularx}
\usepackage{listings}
\usepackage{comment}
\newcommand{\acknowledgments}{\section*{Acknowledgments}}

\usepackage[T1]{fontenc}
\usepackage{textcomp}
\usepackage{microtype}
\usepackage{epstopdf}
\usepackage[hyphens]{url}
\usepackage[bookmarks=false, hidelinks, colorlinks, linkcolor=purple]{hyperref}

\usepackage{enumerate}
\usepackage{multirow}
\usepackage[LGRgreek]{mathastext}
\usepackage{amsmath}

\usepackage{listings} 
\usepackage[dvipsnames]{xcolor}

\DeclareCaptionFormat{custom}{#3} 
\title{\LARGE \bf A Neurosymbolic Approach to Adaptive Feature Extraction in SLAM}

\author{Yasra Chandio$^{1}$, Momin A. Khan$^{1}$, Khotso Selialia$^{1}$, Luis Garcia$^{2}$, Joseph DeGol$^{3}$, and Fatima M. Anwar$^{1}$%
\thanks{$^{1}$Yasra Chandio, Momin A. Khan, Khotso Selialia, and Fatima M. Anwar are affiliated with the University of Massachusetts Amherst, USA. 
}%
\thanks{$^{2}$Luis Garcia is affiliated with the University of Utah, USA.}
\thanks{$^{3}$Joseph DeGol is affiliated with the Steg AI, USA.}
\thanks{Correspondence: ychandio@umass.edu}
}

\begin{document}

\maketitle
\thispagestyle{plain}
\pagestyle{plain}

\begin{abstract}
Autonomous robots, autonomous vehicles, and humans wearing mixed-reality headsets require accurate and reliable tracking services for safety-critical applications in dynamically changing real-world environments.
However, the existing tracking approaches, such as Simultaneous Localization and Mapping (SLAM), do not adapt well to environmental changes and boundary conditions despite extensive manual tuning. 
On the other hand, while deep learning-based approaches can better adapt to environmental changes, they typically demand substantial data for training and often lack flexibility in adapting to new domains. 
To solve this problem, we propose leveraging the neurosymbolic program synthesis approach to construct adaptable SLAM pipelines that integrate the domain knowledge from traditional SLAM approaches while leveraging data to learn complex relationships.  
While the approach can synthesize end-to-end SLAM pipelines, we focus on synthesizing the feature extraction module. 
We first devise a domain-specific language (DSL) that can encapsulate domain knowledge on the essential attributes for feature extraction and the real-world performance of various feature extractors. 
Our neurosymbolic architecture then undertakes adaptive feature extraction, optimizing parameters via learning while employing symbolic reasoning to select the most suitable feature extractor. 
Our evaluations demonstrate that our approach, neurosymbolic Feature EXtraction (\texttt{nFEX}), yields higher-quality features. It also reduces the pose error observed for the state-of-the-art baseline feature extractors ORB and SIFT by up to 90\% and up to 66\%, respectively, thereby enhancing the system's efficiency and adaptability to novel environments.
\end{abstract}


\section{Introduction}
\label{sec:intro}
Accurate tracking is crucial to the wide-scale and practical deployment of autonomous cars, autonomous robots, and mixed-reality applications involving humans~\cite{SLAMAR-klein2007parallel}.  The different agents (e.g., cars, robots, humans) navigate complex and dynamic settings, from busy city streets to cluttered warehouses and ever-evolving homes. To ensure safe and effective operation, the agents require robust tracking mechanisms that can adapt to the constant changes they encounter. 

Traditional Simultaneous Localization and Mapping (SLAM) often struggle in such dynamic scenarios. They rely on pre-defined physical models that may not generalize well to unseen situations or environments with significant lighting variations and scene transitions~\cite{cadena2016past}. Recent advances in artificial intelligence (AI) driven approaches learn non-trivial relationships from data and perform better in dynamic environments and edge cases~\cite{deep-vo-survey-jeong2021comparison}. However, they are data-hungry and lack interpretability~\cite{zhang2021survey-interpretablity}, hindering their deployment in safety-critical applications. Hybrid approaches have aimed to bridge the benefits of data-efficient, physics-based approaches and deep-learning-based approaches, e.g., by preprocessing the input to aid tracking~\cite{Yang_2020_CVPR:D3VO} or placing guard rails around the tracking output to limit the impact of erroneous outcomes~\cite{chen2021dynanet, xing2022slam-dynamic-environmebt-output-profiler}. While effective, the composition of these methods is often ad-hoc and tuned to specific domains or environments and not suited to adaptation~\cite{chen2020survey}.
 
In this work, we formalize these hybrid compositional approaches as neurosymbolic programs~\cite{chaudhuri2021neurosymbolicprogram, ritchie2023neurosymbolic-for-computer-graphics}, which aim to bridge the gap between data-driven learning approaches and rule-based symbolic reasoning. In particular, formulating the SLAM pipeline as a composition of modules--each potentially represented as a neurosymbolic program--enables neurosymbolic program synthesis, where we aim to synthesize tracking programs given a library of neural and symbolic components that fit a dataset and generalize to unseen inputs~\cite{chaudhuri2021neurosymbolicprogram}. While neurosymbolic program synthesis can be employed at the module level or the entire tracking pipeline, we focus on the feature extraction module of the SLAM pipeline as a proof of concept. The feature extraction module detects and tracks feature maps over time. Significant prior work has been on developing feature extraction approaches, such as ORB~\cite{Rublee:2011:ORB} and SIFT~\cite{lowe:2004:SIFT}.
These techniques are well-suited for certain environmental conditions for specific applications~\cite{Karami:2017:Compare, Bansal:2021:2D}. 
Therefore, the feature extraction performance can be improved by dynamically selecting the most appropriate feature extractor and adapting its configuration parameters for the given environmental scenario. 
However, the search space across feature extractors, configurations, agent types, and scenarios is large, with limited performance data available for all combinations.

In this work, we consider finding the right feature extractor and its parameters a neurosymbolic program synthesis task. 
The \emph{neural component} of our neurosymbolic feature extractor,~\texttt{nFEX}, is a standard neural network that finds the optimal parameters for the various feature extractors we consider using a dataset of optimal configurations under different environmental conditions. We empirically generate this dataset by running exhaustive combinations of feature extractors and their parameters through SLAM pipelines. 
\texttt{nFEX}'s \emph{symbolic component} captures the domain knowledge on the attributes of essential features and the impact of environmental conditions on the end-to-end performance of a SLAM pipeline as well as the feature quality metrics, such as texture, dissimilarity, and spatial density.
Moreover, the symbolic representation can incorporate prior knowledge on the end-to-end performance of feature extractors; for example, ORB gives the lowest pose error under bright conditions with good textures.
As with any neurosymbolic program synthesis task, we specify the \emph{symbolic component} using the syntax of a domain-specific language (DSL)~\cite{mernik2005and-DSL}, and we leverage knowledge graphs to represent any symbolic expert knowledge. 
Given these components, we devise a learning algorithm that uses symbolic reasoning to adapt the outcomes of the neural learning component. 
The task for the algorithm is to discover the program’s discrete architecture $\alpha$ (i.e., a feature extractor) and its real-valued parameters $\Theta$ (i.e., feature extractor's configurations).
The task specification that directs this search includes a 
domain-specific quantitative fitness function derived from labeled data or expert knowledge. The algorithm aims to find a program that optimizes the loss under constraints.

The task of neurosymbolic program synthesis is not trivial and requires solving multiple challenges. First, there are no existing SLAM pipelines or feature extraction module DSLs. 
Therefore, we need to devise a DSL to represent various agent types (car, drone, human), their motion types (fast, slow), scene types (indoor, outdoor), light conditions (bright, dark), the set of feature extractors (ORB, SIFT), and their parameters (e.g., no. of features, density), details in \S\ref{sec:fit-func}.  
Second, we must populate a knowledge graph summarizing how different feature extraction approaches perform under various conditions. 
However, as such data is limited in the existing literature~\cite{Karami:2017:Compare, Bansal:2021:2D}, we employed an experiment-driven approach to populate the knowledge graph (details in \S\ref{sec:dsl-framework}). We present our solution to those mentioned above and other challenges in their respective sections. 

We make the following contributions in devising a neurosymbolic program synthesis approach to domain adaptive feature extraction in SLAM pipelines. 
\begin{enumerate}
    \item We develop a DSL for the feature extraction module of a SLAM pipeline that allows expressing the performance of a feature extractor across various scene characteristics. Since limited performance data is available on feature extractors, we leverage an experiment-driven approach to populate the knowledge graph required for the neurosymbolic program synthesis. 
    \item We devise a neural network-based learning approach for feature extractor synthesis that learns deep representations over the set of feature extractors and their parameters. We use a novel two-step fitness function (details in \S\ref{sec:fit-func}) to direct the search toward an optimal feature extractor and its configuration parameters.
    \item We extensively evaluate our proposed approach across three datasets, representing various agent and environmental characteristics and two underlying feature extractors, ORB and SIFT. We demonstrate that our approach dynamically adapts the feature extractor and its parameters. We outperform ORB and SIFT by 90\% and 66\%, using ATE as a metric (details in \S\ref{sec:evaluation}). 
\end{enumerate}

\section{Background \& Related Work}
\label{sec:background}
\subsection{Feature Extraction in SLAM}
\label{sec:feature_extraction_background}
Feature extraction acts as SLAM's eyes, the first module to encounter the raw environmental data. It encodes environmental elements for processing, determining how well it can detect and use landmarks to map and navigate surroundings, influencing the quality of the entire pipeline. 
This process involves operations such as scale-space representation, keypoint detection, transform invariance, and descriptor generation, enabling consistent landmark recognition across observations through techniques such as SIFT and ORB.

First, scale-space representation, utilizing image downsampling and filtering—Gaussian blur for SIFT and image pyramids~\cite{adelson1984pyramid} for ORB—creates a multi-resolution image pyramid, ensuring feature detection across all sizes in the visual field.
Keypoint detection then identifies distinctive locations, such as corners, edges, or blobs, by targeting high-contrast areas using methods like the Harris Corner Detector~\cite{derpanis2004harris-corner}, Difference of Gaussians~\cite{blair2007difference-DoG-scale-space} 
and Hessian matrix~\cite{sasson1973optimal-hessian-matrix} 
to locate keypoints. Transformation invariance enhances this process by normalizing the region around each detected feature. SIFT, for example, determines orientation from local image gradients, rendering descriptors rotation invariant across views. On the other hand, ORB achieves invariance through intensity centroid-based methods. For each detected keypoint, a unique descriptor facilitates feature matching across images, with SIFT employing gradient histograms and a binary descriptor strategy for ORB~\cite{calonder2010brief}.

\subsection{Environmental Influences on Feature Extraction}
\label{sec:environmental_influence_background}
Feature extraction is influenced by environmental changes (light, motion, reflective surfaces, textures), the specific agents navigating it (humans, cars, drones), and the scene's characteristics (indoors or outdoors). This causes several challenges. Lighting is one of several challenges. Too much light can hide image details, while too little makes features hard to detect and match. Motion can blur and displace critical features, complicating their tracking and matching across frames. Reflective surfaces can create false features, affecting map accuracy; well-textured scenes can confuse algorithms, while low-textured environments may offer few features to track, requiring careful selection of distinct features~\cite{yan2022dgs-reflective-texture}.

Furthermore, challenges of agent type and scene, whether navigating high-speed scenarios with cars, adapting to altitude and orientation changes in drones, or enhancing augmented reality experiences in complex indoor settings~\cite{AR-SLAM-survey-cite-billinghurst2015survey}. This customization extends to recognizing features like corners, edges, and textural patterns in indoor environments, tailored to the unique characteristics of man-made structures~\cite{indoor-feature}, while also being adaptable to handle the varying lighting, weather, and natural textures encountered outdoors. Both indoor and outdoor scenes require adjusting to dynamic environmental conditions and the specific demands of more complex environments, such as navigating through low visibility underwater or areas with repetitive manmade patterns. This highlights the need for SLAM systems to adapt to environmental changes and the inherent scene characteristics across different contexts.

Traditional methods struggle with these challenges, affecting feature quality and downstream applications (pose and maps). Adaptive techniques like thresholding and dynamic range adjustment~\cite{yang2018adaptive-dynamic, drago2003adaptive-dynamic-contrast}, optical flow, gyroscopic integration~\cite{hwangbo2009inertial-gyroscopic-motion}, polarization~\cite{zhu2020polarization-fix-reflective} and consistency checks~\cite{ yan2022dgs-reflective-texture}, attempt to mitigate these issues but often fall short in rapidly changing conditions.
However, these methods often fail in rapidly changing conditions primarily because they tend to be ad-hoc solutions or overly specialized to certain environments. Each technique is tailored to address particular issues, such as lighting variations or motion blur, without a holistic understanding of the environment's complexity.  

\subsection{Neurosymbolic Architectures}
\label{sec:neurosymbolic_background}
Recent studies~\cite{ritchie2023neurosymbolic-for-computer-graphics,marconato2024not-neurosymbolic,mao2019neuro-interpreting-scenes} have highlighted the effectiveness of \emph{neurosymbolic program synthesis}~\cite{parisotto2016neuro-program-synthesis} in advancing the capabilities of machines to understand~\cite{kelly2023discovering-nerual-symbolic-iros,mao2019neuro-interpreting-scenes}, navigate~\cite{siyaev2021neuro-speech-understanding}, and interact with their environments~\cite{chen2024enhancing-robot-Program-Synthesis-Environmental-Context}, demonstrating a practical approach to enhancing feature extraction through the integration of learning and reasoning processes.
Neurosymbolic architectures can address the main issues of usual SLAM feature extractors, which can not easily adjust to various environmental changes, affecting their accuracy and reliability. This approach uses the strength of neural networks to process complex data and symbolic systems to apply specific knowledge and rules~\cite{chaudhuri2021neurosymbolicprogram}, making it better suited for different conditions.

Integrating knowledge graphs, DSL, and grammar into SLAM presents a practical realization of neurosymbolic architecture. Knowledge graphs can organize unstructured environmental data and sensor interpretations into a structured format as a dynamic reference point. Using this graph, a DSL can provide a framework for expressing operations and configurations of feature extraction that align with the domain's specific requirements. To ensure coherence and semantic integrity, the grammar of a DSL can dictate the rules that structure the language, defining how symbols, keywords, and operators combine to form valid and meaningful expressions.
This approach can fine-tune feature extraction parameters for challenges like low-light visibility and navigating reflective surfaces and select the best extractor for the situation. For instance, by symbolically analyzing motion and texture, it can learn to adjust motion compensation and feature quantity, steering towards extractors ideal for varied textures. 

\begin{figure}[t]
    \centering
    \includegraphics[width=\columnwidth]{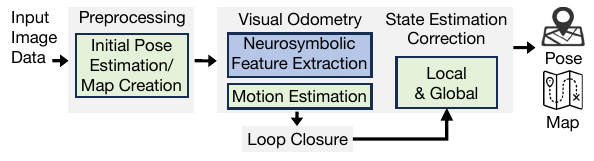}
    \vspace{-0.7cm}
    \caption{\emph{Our contribution (blue box) within SLAM pipeline.}}
    \label{fig:highlevel-overview}
    \vspace{-0.1cm}
\end{figure}

\begin{figure}[t]
    \centering
    \includegraphics[width=\linewidth]{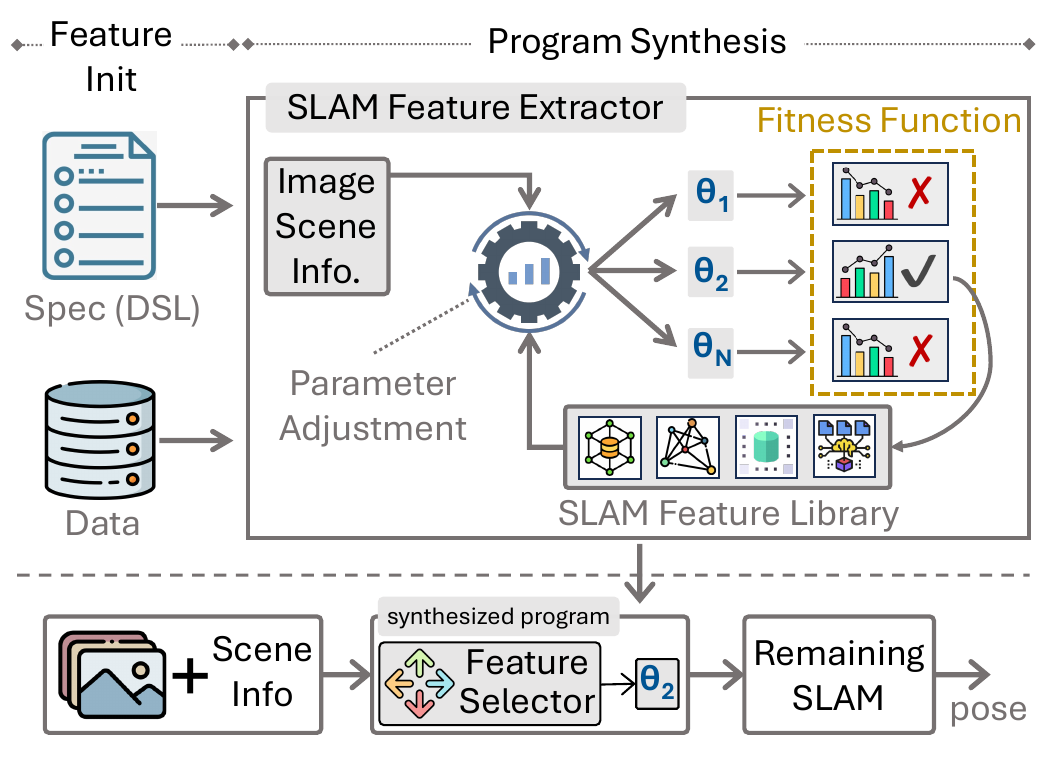}
    \caption{\emph{A high-level overview of our approach.}
    }
    \label{fig:overview}
    \vspace{-0.1cm}
\end{figure}

\begin{figure*}[t]
    \centering
    \includegraphics[width=.95\linewidth]{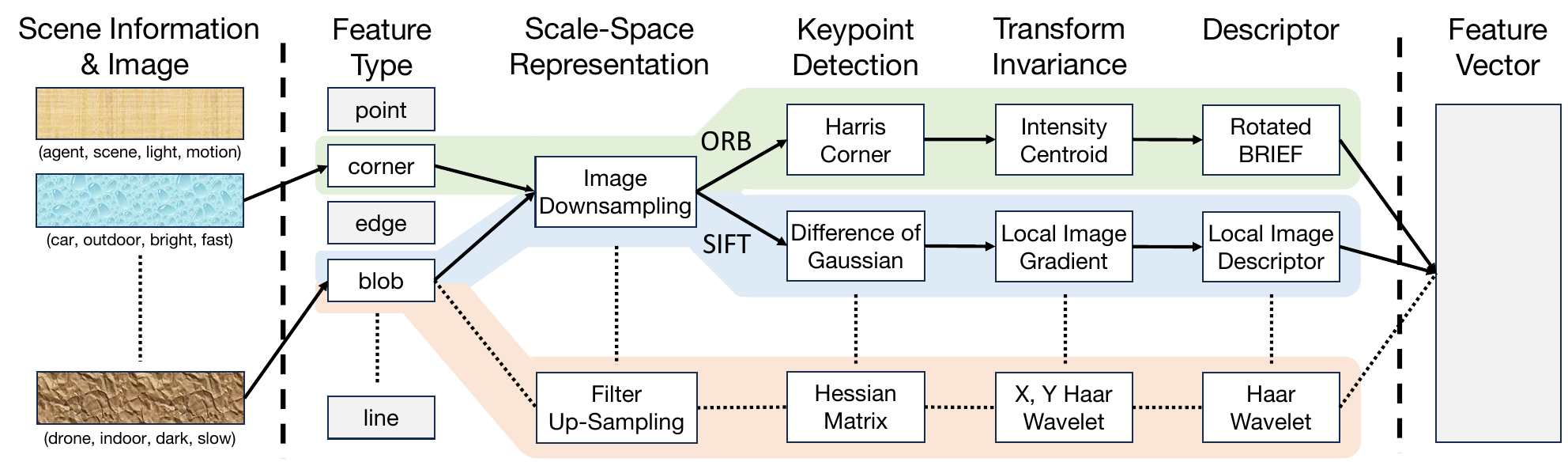} 
    \vspace{-0.15cm}
    \caption{\emph{Illustration of feature extraction knowledge graph.}}
     \vspace{-0.25cm}
    \label{fig:knowledge-graph}
\end{figure*}

\section{DSL Construction for Neurosymbolic SLAM}
\label{sec:method}
This section details \texttt{nFEX} approach, leveraging neurosymbolic programming to adapt feature extraction dynamically.

\subsection{Neurosymbolic Program}
\label{sec:program-synthesis}
In traditional SLAM, images are processed by a pre-selected feature extractor with fixed parameters, producing a feature vector fed to the remaining SLAM pipeline (odometry and other optimizations) for pose generation. To this end, our proposed approach \texttt{nFEX} illustrated in Figure \ref{fig:overview}, synthesizes a neurosymbolic program designed to dynamically select and configure feature extractors based on real-time environmental inputs by replacing the traditionally manually-tuned feature extraction module with our \emph{neurosymbolic feature extraction}, highlighted in blue color in Figure \ref{fig:highlevel-overview}. 

\subsubsection{DSL Framework}
\label{sec:dsl-framework}
This process begins with establishing a DSL framework, which outlines the structure for: 

\noindent \emph{Feature Extractor Selection ($\alpha$)} identifies the optimal feature extractor architecture for the given scene conditions.

\noindent \emph{Parameter Configuration ($\Theta$)} determines the best parameter settings to enhance feature extractor performance.

It encapsulates domain insights into the feature extraction module with a knowledge graph that acts as a database of domain information that helps understand and navigate the module's operation and parameters, as illustrated in Figure \ref{fig:knowledge-graph}; we generate this graph as one of our contributions. It encompasses nodes representing various feature extraction operations, such as scale space representation, keypoint detection, and descriptor. Edges within the graph delineate the relationships and compatibilities between these operations, informed by performance metrics, computational considerations, and adaptability to environmental factors. 
This graph enriches both our synthesis and the DSL's capacity to incorporate novel feature extraction methods in the future. 

\begin{figure}[ht!]
    \centering
    \includegraphics[width=\linewidth]{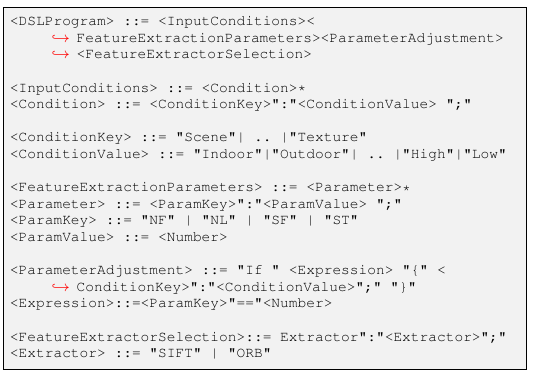}
    \caption{Grammar to parse the DSL for \texttt{nFEX}.}
    \label{figure:DSL-grammer}
\end{figure}

Next, \emph{grammar}, outlined in Figure ~\ref{figure:DSL-grammer}, provides a blueprint for creating the DSL program that can interpret and act upon input conditions (environmental context). Using the graph, the grammar parses the structure for input conditions (such as indoor/outdoor scene, agent types, lighting conditions, and so on), and each rule describes a part of the program for selecting the suitable $\alpha$ and their $\Theta$. Our DSL's grammar enhances adaptability and streamlines each recalibration for new applications, making it straightforward for users to add new input conditions and parameters.

This DSL ensures the synthesized programs are logically coherent and actionable, allowing for automated parsing and execution of feature extraction. An example of one such program is illustrated in Figure \ref{figure:DSL-program}. Once the DSL is established, we operationalize it through a two-phase process:

\begin{figure}[ht!]
    \centering
    \includegraphics[width=\linewidth]{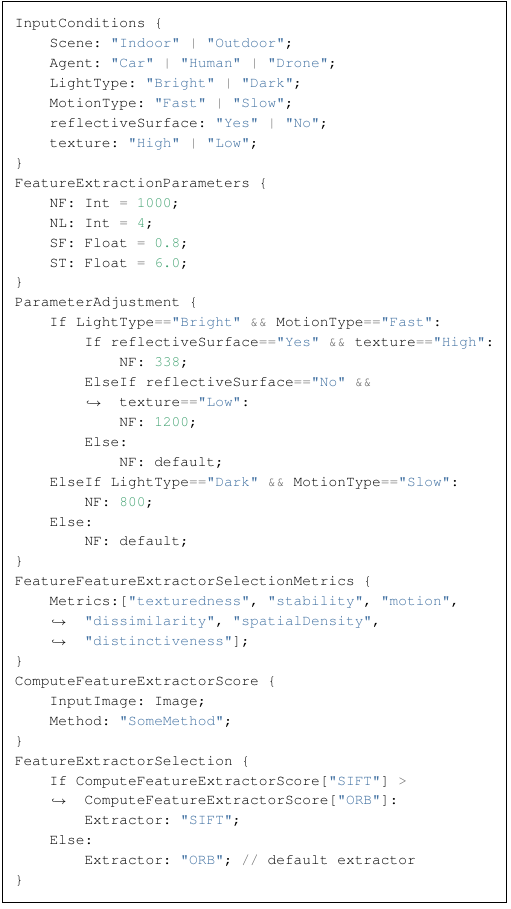}
    \caption{Example DSL program}
    \label{figure:DSL-program}
\end{figure}
This DSL ensures the synthesized programs are logically coherent and actionable, allowing for automated parsing and execution of feature extraction. An example of one such program is illustrated in Figure \ref{figure:DSL-program}. Once the DSL is established, we operationalize it through a two-phase process:

\subsubsection{Program Synthesis (Training)}
\label{sec:program-training}
We first optimize feature extractor parameters ($\Theta$) using neural networks (details in \S~\ref{sec:fit-func}) to generalize the extractor's performance across different contexts. To handle the inherent uncertainties and partial unstructured information in real-world scenarios, we employ smooth interpretation~\cite{smooth-interpretation} techniques, allowing for a nuanced adaptation. Then, with this optimized $\Theta$, the program evaluates to make informed decisions about the most suitable $\alpha$ and its configuration for the given context. 
This selection process is akin to neural architecture search ~\cite{elsken2019neural-NAS}, where various architectures (feature extractors in this context) are evaluated for contextual performance. Finally, we employ the feedback mechanism integral to this phase, allowing continuous program refinement based on downstream performance (e.g., pose). This iterative process ensures the program remains optimally configured to adapt to new challenges and environments.
\subsubsection{Program Use (Inference)}
\label{sec:program-use}

During inference, our synthesized program dynamically applies optimized $\Theta$) and $\alpha$ in real-time scenarios. It dynamically adjusts to changing environments, analyzing scene data for accurate feature extraction and integration into the SLAM pipeline. This phase emphasizes real-time adaptation and continuous learning from new data, enabling ongoing system performance refinement and boosting adaptability across diverse scenarios.

\section{Implementation}
\label{sec:fit-func}
\subsection{Fitness Function for DSL formulation}
We implement a DSL program through a two-step fitness function to materialize our approach.

\subsubsection{Parameter Tuning ($\Theta$)}
Given a set of feature extractors $\alpha = \{\alpha_1, \alpha_2, \ldots, \alpha_y\}$, initial parameters \(\Theta = \{\theta_1, \theta_2, \ldots, \theta_n\}\) for each feature extractor and a set of \emph{symbolic} environmental conditions \(E = \{e_1, e_2, \ldots, e_c\}\) the optimization process is defined by a transformation function \(g(\theta, E)\), which \emph{dynamically adjusts} \(\Theta\) based on \(E\), resulting in an optimized parameter set $\Theta'= \{\theta_1', \theta_2', \ldots, \theta_n'\}$ with \(w_k\), the weight for each parameter \(k\):
\vspace{-0.3cm}
\[
\Theta' = g(\Theta, E), \quad w_k = \prod_{j=1}^{c} f_{kj}(e_j)
\]
\[
\theta_k' = w_k \cdot \theta_k = \left( \prod_{j=1}^{c} f_{kj}(e_j) \right) \cdot \theta_k \quad \text{for each } k \in \{1, 2, \ldots n\}
\]
The term $\prod_{j=1}^{c} f_{kj}(e_j)$ is the cumulative product of adjustments from all the environmental conditions for a given parameter $k$. Finally, the fitness of the optimized parameters \(\theta'\) in the context of environmental conditions \(E\) is then evaluated by the fitness function \(F\):
\[
F(\Theta' | E) = f(\Theta' | E)
\]
This formulation demonstrates how scene conditions directly influence parameter optimization, allowing for adaptable feature extraction based on the environment.

\subsubsection{Feature Extractor Selection ($\alpha$)}
\label{sec:fitness_quality_evaluation}
Once we obtain $\Theta'$ for each feature extractor \(\alpha_i\), given set of \emph{symbolic} metrics $M = {m_1, m_2 \ldots m_p}$ with dynamically adjusted weights $\rho_x = h_x(E)$ for each $x \in \{1, 2, \ldots, p\}$. \(h_x(E)\) is a function that computes the weight of the \(x^{th}\) metric based on the environmental conditions $E$. We assess the quality of detected features for each feature extractor \(\alpha_i\) with its parameters optimized as \(\Theta_i'\):
\[
F_{\text{metrics}}(\alpha_i | \Theta_i', E) =  \sum_{x=1}^{p} \rho_x  \cdot m_x(\alpha_i') = \sum_{x=1}^{p} h_x(E) \cdot m_x(\alpha_i')
\]
The final selection of the feature extractor \(\alpha^*\) that maximizes \(F_{\text{metrics}}\):
\[
\alpha^* = \underset{\alpha_i \in \{\alpha_1, \alpha_2, \ldots, \alpha_y\}}{\arg\max} \, F_{\text{metrics}}(\alpha_i | \Theta_i', E)
\]
This ensures that the feature extractor's selection and performance evaluation are adapted to the environmental context.

\subsection{Experimental Setup}
\subsubsection{Feature Parameters ($\Theta$)}
Our approach prioritizes optimizing parameters that influence feature extraction operations and are responsive to our defined environmental inputs. To this end, we choose the following parameters:

\noindent \emph{\textbf{Number of Features}} \textbf{($\text{NF}$)} affects the system's ability to recognize and track environmental landmarks. Optimizing $\text{NF}$ is crucial for balancing detail capture with computational load, ensuring enough features are detected to represent the scene accurately without overwhelming the processing capacity, particularly in complex or sparse environments.

\noindent \emph{\textbf{Scale Factor }} \textbf{($\text{SF}$)} determines the interval between successive scales in the image pyramid, influencing the system's capacity for scale-invariant feature detection. A properly tuned $\text{SF}$ allows the feature extractor to consistently identify features regardless of their size in the image, enabling the system to remain robust to changes in the size or distance of objects affected by lighting conditions and motion.

\noindent \emph{\textbf{Number of Pyramid Levels }} \textbf{($\text{NL}$)} adjusts for features across object sizes. It expands the system's capability to discern features of various sizes across the entire scene, directly impacting the system's versatility in handling scenes with diverse spatial characteristics. 

\noindent \emph{\textbf{Keypoint Selectivity Thresholds}} \textbf{($\text{ST}$)}, ensure distinct, reliable keypoints for matching across images, crucial for dealing with reflective surfaces or inconsistent textures by promoting uniform feature distribution.
\subsubsection{Feature Extractors ($\alpha$)}
We integrated our neurosymbolic feature extraction module into ORBSLAM3~\cite{ORBSLAM3_TRO} due to its advanced adaptability and modular design, enabling effective evaluation of our \texttt{nFEX} approach within its framework. By selecting both ORB for its computational efficiency in corner detection and SIFT for its precision in blob detection, we aim to assess the distinct advantages each feature extractor brings to the SLAM. This approach allows for a comprehensive comparison of corner-based and blob-based feature extraction methods, leveraging our fitness functions to optimize and evaluate their performance.

\subsubsection{Feature quality metrics ($M$)}
Our choice of $M$ focuses on metrics intrinsic to feature extraction, allowing our neurosymbolic optimization to influence them without relying on downstream SLAM data.

\noindent \textbf{Texturedness} (\(m_1 = \sqrt{\sum (I(a, b) - \bar{I})^2}\)) integrates texture variance within the image
    with \(I(a, b)\) being pixel intensity and \(\bar{I}\) the average intensity.
    
    \noindent \textbf{Dissimilarity} (\(m_2= \sum |I_{\text{feature}} - I_{\text{surrounding}}|\)) measures feature uniqueness against its surroundings.
    
    \noindent \textbf{Motion} (\(m_3= \| \mathbf{a}_{t+1} - \mathbf{a}_t \|\)) quantifies feature movement between frames.

    \noindent \textbf{Stability} (\(m_4 = \frac{\text{Number of stable features}}{\text{Total number of features}}\)) evaluates detection consistency over time.    
    
    \noindent \textbf{Spatial density} (\(m_5 = \frac{\text{Number of Features}}{\text{Area of Region}}\)) examines evenness of feature distribution.

    \noindent \textbf{Distinctiveness} (\(m_6= \frac{1}{\sum \text{similarity(feature, other features)}}\)) gauges how features stand out from each other.

    \noindent \textbf{Repeatability} (\(m_7 = \frac{\text{Number of Re-detected Features}}{\text{Total Number of Features}}\)) measures consistent feature detection.
\begin{table*}[t]
\centering
\vspace{0.1cm}
\caption{Performance comparison of different feature extractors using Mean ATE in meters (averaged over 10 runs).}
\label{tab:end-to-end}
\footnotesize
\begin{tabular}{|cc|cc|cc|cc|}
\hline
\multicolumn{2}{|c|}{\multirow{2}{*}{}}               & \multicolumn{2}{c|}{EuRoC} & \multicolumn{2}{c|}{KITTI}  & \multicolumn{2}{c|}{HoloSet} \\ \cline{3-8} 
\multicolumn{2}{|c|}{} & \multicolumn{1}{c|}{MH01} & MH05 & \multicolumn{1}{c|}{KITTI-1} & KITTI-6 & \multicolumn{1}{c|}{Campus-Center-1} & Suburb-Jog-2 \\ \hline \hline

\multicolumn{1}{|c|}{\multirow{2}{*}{ORB}}  & Default & \multicolumn{1}{c|}{0.855} & 0.952 & \multicolumn{1}{c|}{2.955} & 1.173  & \multicolumn{1}{c|}{11.789}  & 11.604  \\ \cline{2-8} 

\multicolumn{1}{|c|}{}                      & Dynamic & \multicolumn{1}{c|}{0.792} & 0.815 & \multicolumn{1}{c|}{\textbf{0.565}} & 0.126 & \multicolumn{1}{c|}{5.903} & 5.845 \\ \hline

\multicolumn{1}{|c|}{\multirow{2}{*}{SIFT}} & Default & \multicolumn{1}{c|}{0.860} & 1.038 & \multicolumn{1}{c|}{\emph{fail}} & \emph{fail}  & \multicolumn{1}{c|}{13.789}  & 12.825  \\ \cline{2-8} 

\multicolumn{1}{|c|}{}    & Dynamic & \multicolumn{1}{c|}{0.859} & 0.882 & \multicolumn{1}{c|}{6.426} & 5.875  & \multicolumn{1}{c|}{7.049}  & 6.984  \\ \hline\hline
\multicolumn{2}{|c|}{nFEX}                            & \multicolumn{1}{c|}{\textbf{0.704}} & \textbf{0.761} & \multicolumn{1}{c|}{\textbf{0.565}} & \textbf{0.115}  & \multicolumn{1}{c|}{\textbf{4.729}}  & \textbf{5.800}  \\ \hline\hline
\end{tabular}
\end{table*}
\subsubsection{Datasets}
We use three datasets to evaluate adaptability across various agents, scenes, and motion types: KITTI~\cite{KITTI-geiger2013vision}, for urban outdoor scenes with fast-moving cars; EuRoC~\cite{burri2016euroc}, focusing on diverse indoor environments navigated by drones with dynamic motions; and HoloSet~\cite{chandio2022holoset}, covering both indoor and outdoor settings from a human perspective, capturing natural human motions.

Our approach requires scene attributes to adapt dynamically to diverse environments. Because they significantly impact feature extraction, we selected specific scene condition labels, such as indoor/outdoor settings, agent types, lighting conditions, motion speeds, reflective surfaces, and texture levels.
We derived these labels through methods proposed in \cite{dataset-characterization-ali2022we} and corroborated with scene information from ~\cite{guo2022lidar-slam-featrue-and-KITTI-scene-info},~\cite{burri2016euroc}, and ~\cite{chandio2022holoset}.
However, we acknowledge that creating exhaustive verification mechanisms for dataset characterization against real-world scenarios is an open problem and beyond the scope of this paper.

\subsubsection{Fitness Function Training}
We use a two-step neural network to optimize feature extraction parameters \( F(\Theta'|E) \) and feature extractor selection \( \alpha^* \). \( F(\Theta'|E) \) training uses a Multi-Layer Perceptron (MLP) fully connected layers ($\times 3$) and ReLU activations ($\times 2$), minimizing the loss as:
\[
\mathcal{L}(\Theta) = \frac{1}{N} \sum_{i=1}^{N} (y_i - \hat{y}_i(\Theta))^2
\]
where \( N \) is the batch size, \( y_i \) the true label, and \( \hat{y}_i(\Theta) \) the predicted value. 
Next, \( \alpha^* \) training uses a hybrid neural network combining Convolutional Neural Network (CNN) layers for image analysis and fully connected layers ($\times 3$) for numerical data, refining feature quality by minimizing the cross-entropy loss, which quantifies the discrepancy between predicted probabilities \( \hat{l}_i(\alpha^*) \) and true labels \( l_i \) as:
\[\mathcal{L}(\alpha^*) = -\frac{1}{N} \sum_{i=1}^{N} \left[ l_i \log \hat{l}_i(\alpha^*) + (1 - l_i) \log (1 - \hat{l}_i(\alpha^*)) \right]\]
Feature scaling is applied using a standard scaler to ensure features have a mean of zero and a standard deviation of one, making them more suitable for neural network training.

\section{Evaluation}
\label{sec:evaluation}
We evaluate the integration of \texttt{nFEX} with ORBSLAM3 by comparing performance using the Absolute Trajectory Error (ATE). It calculates the deviation between aligned estimated \(T_{est}\) and ground truth trajectories \( T_{gt} \) as: $ATE = \sqrt{\frac{1}{N} \sum_{i=1}^{N} || T_{gt}(i) - T_{est}(i) ||^2}$.
Here $N$ represents the total number of trajectory points being compared between the $T_{est}$ and $T_{gt}$. 
Though precise trajectories do not guarantee flawless maps, they are widely used to gauge SLAM performance.
We use ORB and SIFT with default (parameters from OpenCV) and dynamically optimized parameters ($\Theta$) as baselines, and we trained our models on sequences EuroC-MH\_02, KITTI-00, and Holoset-Suburb-seq-1.
Lastly, we ran all our tests on NVIDIA GeForce RTX 2070 GPU, with a batch size of 8, Adam optimizer, and a learning rate of $1e^{-4}$.

\subsection{End-to-End Performance}
\label{sec:illus-exp}
Table \ref{tab:end-to-end} showcases the Mean ATE for ORB and SIFT feature extractors with default and dynamically optimized parameters alongside our \texttt{nFEX} approach across sequences from all three datasets. 
We can see that dynamic optimization significantly enhances ORB's performance, notably in the KITTI sequences, where it drastically lowers Mean ATE, underscoring the importance of parameter tuning. However, SIFT fails in its default mode for KITTI and shows limited improvement even when dynamically optimized.

In challenging indoor scenarios of EuRoC and mixed environments of HoloSet, \texttt{nFEX} achieves the lowest Mean ATE, highlighting its effectiveness in complex settings. Even after optimization, SIFT's failure in KITTI's default configuration and its lagging performance also highlights \texttt{nFEX}'s edge in achieving a balance between robust feature extraction and computational efficiency.
Overall, \texttt{nFEX} shows its potential to significantly enhance SLAM by adapting to the operational environment, choosing the best feature extractor, and optimizing parameters dynamically.

\subsection{Generalization}
\label{sec:eval_generalization}
Table \ref{tab:generalization} highlights the generalization capability of \texttt{nFEX}. In the first case, MH01 -- MH01, the training on the first 70\% of the sequence and testing on the remaining 30\% of the sequence
yielded a mean ATE of 0.792m, demonstrating \texttt{nFEX}'s effectiveness within familiar environments. When tested on MH05 (difficult sequence) after training on MH01 (easy sequence), the ATE increased to 1.329m, indicating a reasonable generalization to a different sequence within the same dataset despite new environmental and motion dynamics. The more significant jump to 2.476m when transitioning from training on MH01 to testing on KITTI-6 reflects the challenge of adapting to entirely different environmental conditions, such as outdoor versus indoor settings and drone versus car motion dynamics.
These results underscore \texttt{nFEX}'s potential for cross-environment generalization.

\begin{table}[]
\centering
\footnotesize
\caption{Generalization performance of \texttt{nFEX} when trained on the first (or part of a) sequence and tested on the second sequence in the column heading. 
}
\label{tab:generalization}
\resizebox{\columnwidth}{!}{%
\begin{tabular}{|cl|c|c|c|}
\hline
\multicolumn{2}{|c|}{}     & MH01--MH01 & MH01--MH05 & MH01 -- KITTI-6 \\ \hline
\multicolumn{2}{|c|}{nFEX} & 0.792 m                        & 1.329 m                        & 2.476 m                               \\ \hline
\end{tabular}%
}
\end{table}

\subsection{Ablation Study}
\label{sec:eval_ablation}
\subsubsection{Impact of Parameter Optimization}
\begin{figure}
    \centering   \includegraphics[width=.95\columnwidth]{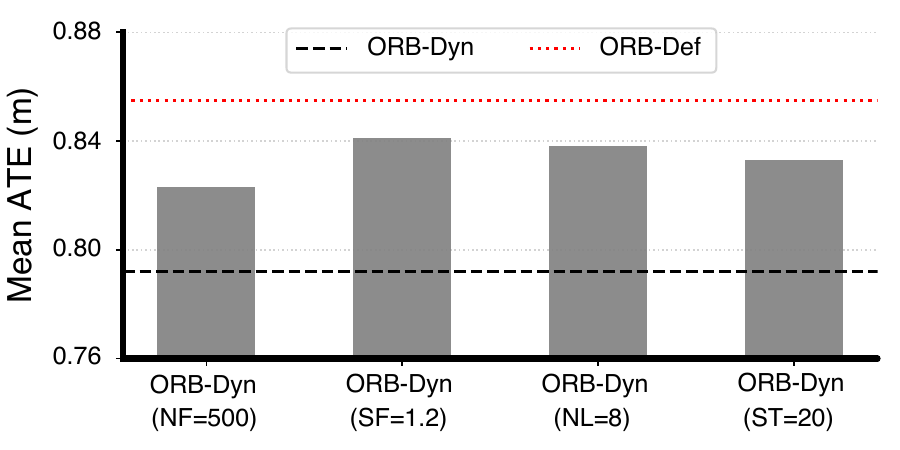}
    \vspace{-0.25cm}
    \caption{Mean ATE in meters for ORB-Dyn where one parameter is fixed at a time. 
    }
    \label{fig:f1-ablation}
    \vspace{-0.1cm}
\end{figure}
In Figure \ref{fig:f1-ablation}, we examine the effect of individual parameters— $NF, SF, NL, ST$—on the ATE for the ORB feature extractor. We fixed one parameter and kept the other dynamic. ORB with all dynamic or all default parameters are highlighted by black and red lines, respectively. This comparison shows the potential of dynamic parameter configurations, particularly in complex and varied environments. 
Further analysis shows \texttt{nFEX} and dynamically optimized ORB (ORB-Dyn) significantly increase feature matches over default settings (Table \ref{tab:num_matches}). \texttt{nFEX} notably outperforms default ORB (ORB-Def) and SIFT with dynamic parameters (SIFT-Dyn), showcasing its ability to optimize feature extraction for superior match quality adaptively.

\begin{table}[t]
\centering
\footnotesize
\caption{Average number of feature matches for different feature extractors for MH01 sequence.
}
\vspace{-0.15cm}
\label{tab:num_matches}
\resizebox{\columnwidth}{!}{%
\begin{tabular}{|cl|c|c|c|c|c|}
\hline
\multicolumn{2}{|c|}{}     & ORB-Def & ORB-Dyn & SIFT-Def & SIFT-Dyn & \texttt{nFEX} \\ \hline
\multicolumn{2}{|c|}{\# of features} & 500 & 418 & no limit  &   518  &  529       \\ \hline
\multicolumn{2}{|c|}{\# of matches} & 389 & 602 & 1101  &   328  &  772       \\ \hline
\end{tabular}%
}
\end{table}
\subsubsection{Impact of Feature Extractor Selection}
\begin{table}[t]
\centering
\footnotesize
\caption{Selection of different feature extractors by \texttt{nFEX} for different frames within MH01 sequence.
}
\vspace{-0.15cm}
\label{tab:ext_selection}
\resizebox{\columnwidth}{!}{%
\begin{tabular}{|cl|c|c|c|c|}
\hline
\multicolumn{2}{|c|}{}     & ORB-Def & ORB-Dyn & SIFT-Def & SIFT-Dyn\\ \hline
\multicolumn{2}{|c|}{\# of frames} & 613 / 3638 & 1366 / 3638 & 535 / 3638  &   1124 / 3638           \\ \hline
\end{tabular}%
}
\end{table}
Table \ref{tab:ext_selection} illustrates \texttt{nFEX}'s feature extractor selection on the $MH01$ sequence, which dynamically switches between ORB and SIFT extractors. The selection process results in a diverse distribution: ORB-Dyn is selected for 1366 out of 3638 frames, demonstrating its preference under specific conditions, followed by SIFT-Dyn with 1124 frames. In contrast, ORB-Def and SIFT-Def (SIFT with default parameters) are chosen less frequently, indicating that dynamic parameter optimization generally outperforms default settings.
\subsection{Computation}
\label{sec:eval_computation}
In Table \ref{tab:timing}, we show the computational efficiency of \texttt{nFEX} averaged over 10 runs. Model loading takes 1800 milliseconds (ms), a one-time overhead. 
Parameter and feature extractor selection are executed swiftly, ensuring rapid adaptability. At 112 ms, frame processing remains within acceptable bounds for real-time application, positioning \texttt{nFEX} as a viable solution for enhancing SLAM performance with negligible impact on processing speed.
\begin{table}[t]
\centering
\footnotesize
\caption{Timing for tasks in \texttt{nFEX}}
\vspace{-0.15cm}
\label{tab:timing}
\resizebox{\columnwidth}{!}{%
\begin{tabular}{|cl|c|c|c|}
\hline
\multicolumn{2}{|c|}{}      & Parameter Selection & Frame Processing & Extractor Selection \\ \hline
\multicolumn{2}{|c|}{Time} &  5.49 $\text{$\mu$}$s & 112 ms  &  4.37 $\text{$\mu$}$s       \\ \hline
\end{tabular}%
}
\end{table}

\section{Limitations and Future Work}
\label{sec:discussion}
\noindent \textbf{Data efficiency.}
Demonstrating significant data efficiency, \texttt{nFEX} aligns with prior work~\cite{YI:2018:NSVQA}, suggesting a minimal data subset can yield substantial system efficiency. Enhancing this efficiency further will make \texttt{nFEX} adaptable to diverse applications, particularly where data collection is challenging.

\noindent \textbf{Interpretability.} \texttt{nFEX}'s dynamic extractor selection for varying scenes enhances model transparency to a certain degree. By analyzing time series data, we can illustrate how and why certain extractors are chosen, offering insights based on logical reasoning derived from the context. This is crucial for critical applications where understanding the model's decision-making process is as important as its performance. Moreover, incorporating physical models to offer performance guarantees akin to control systems will enhance \texttt{nFEX}'s reliability and predictability in safety-critical applications such as autonomous driving.

\noindent \textbf{Improved cross-environment generalization.} While promising within similar datasets, \texttt{nFEX} faces some challenges across different environments, necessitating dataset-specific tuning for optimal performance. Addressing this will further our goal of universal SLAM adaptability.

\noindent \textbf{Resource Optimization} Tailoring \texttt{nFEX} to operate within the constraints of specific resources, as informed by the agent type (e.g., computational power available on a drone versus a car), is an area for improvement. Optimizing the program to utilize available resources efficiently can enhance deployment flexibility and operational efficiency, ensuring \texttt{nFEX} is effective across a wide range of use cases. \texttt{nFEX}'s neurosymbolic architecture is well-suited for platform-aware neurosymbolic architecture search approaches, e.g., ~\cite{saha2023tinyns}.

This work is the first step toward leveraging neurosymbolic programming for domain-adaptive SLAM. It guides the community toward tackling the outlined challenges and exploring open problems. By continuing to refine and build upon these concepts, we aim to steer future research toward developing more adaptable, efficient, and explainable SLAM.

\section{Conclusion}
\label{sec:conclusion}
Accurate and adaptive tracking remains a key challenge for achieving agents' safe and robust operation in dynamic environments. Our approach leverages the strengths of both symbolic reasoning and data-driven learning to dynamically select and configure the most appropriate feature extractor based on the encountered environment. We demonstrated how nFEX's dynamicity significantly improves performance across three representative benchmark datasets in different domains compared to traditional approaches. While this work focuses on adaptive SLAM feature extraction, the modular nature of the SLAM pipeline paves the way for extending neurosymbolic program synthesis to other modules as well.

\acknowledgments{%
The research reported in this paper was sponsored in part by the National Science Foundation (NSF) under awards 2435642 and 2237485. The views and conclusions contained in this document are those of the authors and should not be interpreted as representing the official policies, either expressed or implied, of the funding agencies.
}

\bibliographystyle{ieeetr}

\end{document}